\documentclass{article}

\pdfoutput=1

\usepackage{arxiv}

\usepackage[utf8]{inputenc} 
\usepackage[T1]{fontenc}    
\usepackage{hyperref}       
\usepackage{url}            
\usepackage{booktabs}       
\usepackage{amsfonts}       
\usepackage{nicefrac}       
\usepackage{microtype}      
\usepackage{lipsum}		
\usepackage{graphicx}
\usepackage{natbib}
\usepackage{doi}
\usepackage{amsmath}
\usepackage{algorithm}
\usepackage{algpseudocode}

\title{Exploring the In-context Learning Ability of Large Language Model for Biomedical Concept Linking}

\author{Qinyong Wang \\
    Center for Artificial Intelligence in Drug Discovery\\
    Case Western Reserve University\\
    Cleveland, OH, 44106 \\
    \texttt{qxw225@case.edu} \\
    \And
    Zhenxiang Gao \\
    Center for Artificial Intelligence in Drug Discovery\\
    Case Western Reserve University\\
    Cleveland, OH, 44106 \\
    \texttt{zxg306@case.edu} \\
    \And
    Rong Xu \\
    Center for Artificial Intelligence in Drug Discovery\\
    Case Western Reserve University\\
    Cleveland, OH, 44106 \\
    \texttt{rxx@case.edu} \\
}

\date{}

\hypersetup{
pdftitle={A template for the arxiv style},
}

\begin{document}
\maketitle
\begin{abstract}
The biomedical field relies heavily on concept linking in various areas such as literature mining, graph alignment, information retrieval, question-answering, data, and knowledge integration. Although large language models (LLMs) have made significant strides in many natural language processing tasks, their effectiveness in biomedical concept mapping is yet to be fully explored. This research investigates a method that exploits the in-context learning (ICL) capabilities of large models for biomedical concept linking. The proposed approach adopts a two-stage retrieve-and-rank framework. Initially, biomedical concepts are embedded using language models, and then embedding similarity is utilized to retrieve the top candidates. These candidates' contextual information is subsequently incorporated into the prompt and processed by a large language model to re-rank the concepts. This approach achieved an accuracy of 90.1\% in BC5CDR disease entity normalization and 94.7\% in chemical entity normalization, exhibiting a competitive performance relative to supervised learning methods. Further, it showed a significant improvement, with an over 20-point absolute increase in F1 score on an oncology matching dataset. Extensive qualitative assessments were conducted, and the benefits and potential shortcomings of using large language models within the biomedical domain were discussed.
\end{abstract}


\section{Introduction}
\paragraph{}
Biomedical concept linking is a critical procedure in knowledge integration\cite{himmelstein2017systematic} and information retrieval\cite{gopalakrishnan2019survey}. This process identifies biomedical concepts within the text and associates these concepts with matching entities in a biomedical knowledge base. It essentially forms a bridge between text and structured knowledge databases, facilitating the efficient extraction and utilization of intricate biomedical information. Concept linking is integral to diverse applications, such as literature mining, graph alignment\cite{zeng2021comprehensive}, and information retrieval within the biomedical domain. Moreover, the efficacy of concept linking directly influences the performance of graph-based algorithms, search algorithms, and question-answering systems.

\paragraph{}
While significant progress has been made in the field of biomedical concept linking, major issues still pertain to their limited capacity to handle the ambiguity and complexity characteristic of biomedical concepts\cite{huang2020biomedical}. Supervised Training or fine-tuning methods typically require extensive labeled data, which is labor-intensive and expensive to compile \cite{ji2020bert,li2019fine}. The reliance on labeled data also introduces the data expiration problem, given the evolving nature of biomedical knowledge. For instance, biomedical entities linking datasets often utilize an ontology system for labeling text mentions; however, these systems change over time. The Medical Subject Headings (MeSH) system housed 28,000 concepts in 2016, but by 2023, the count has increased to 32,000 concepts\cite{lipscomb2000medical}. Consequently, if we were to employ label-dependent supervised methods, models may need recurrent retraining to stay updated. Additionally, these techniques are often task-specific, lacking the requisite adaptability to handle different datasets or tasks without comprehensive retraining.

\begin{figure}[h]
  \centering
  \includegraphics[scale=0.2]{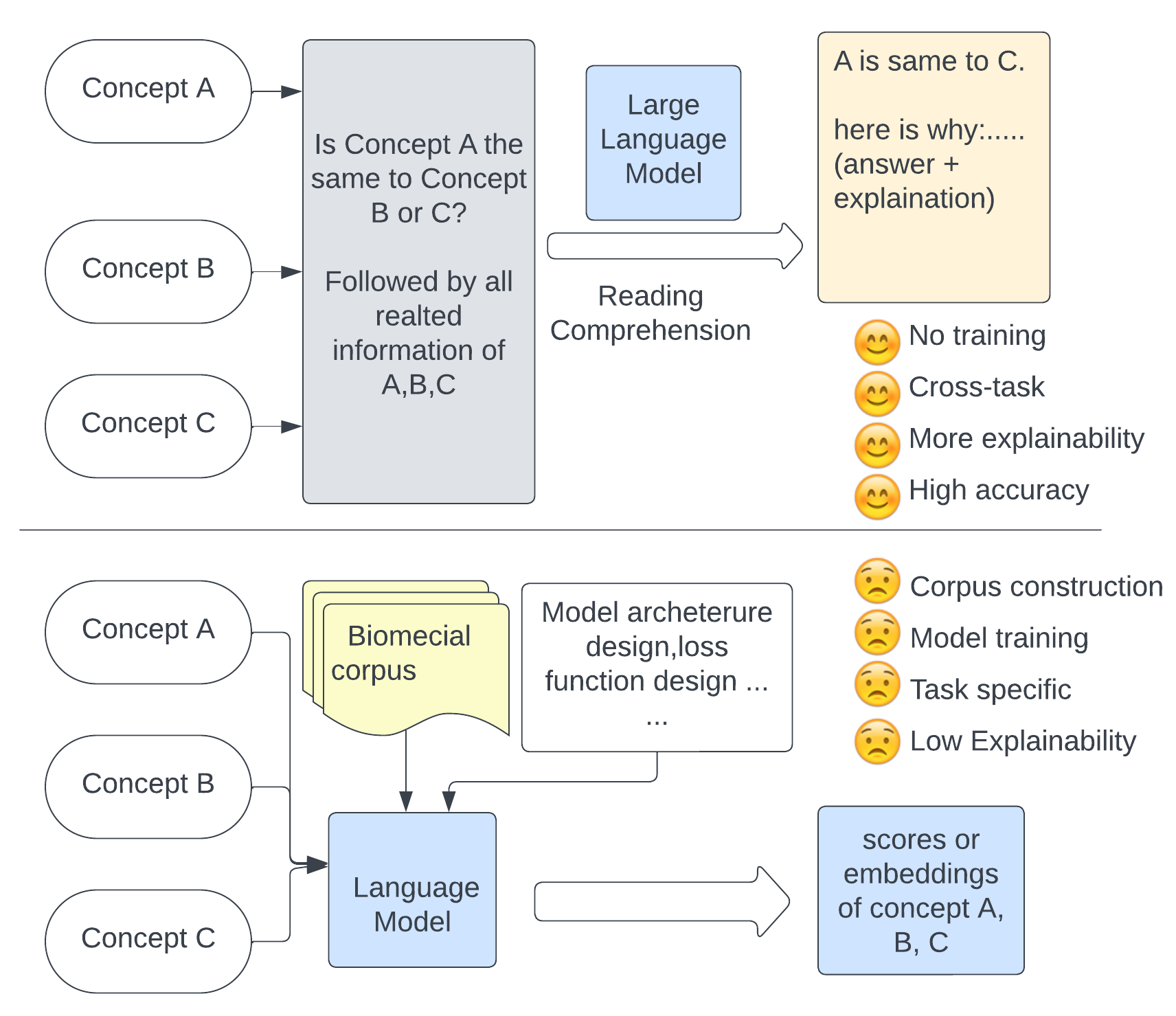}
  \caption{Overview of the proposed methodology and comparison with previous methods.}
  \label{fig:methodology1}
\end{figure}

\paragraph{}
There is a need for a more generalized framework for biomedical concept linking that can navigate the complex landscape of biomedical text data effectively. The ideal system should be adaptable, capable of processing various datasets and tasks without the need for task-specific training data. It should also possess the robustness to handle the ambiguity and complexity of biomedical concepts. Biomedical concept linking can be described as the process of identifying concepts within a given text and associating them with corresponding concepts in a biomedical knowledge base. This task encompasses a range of specific tasks, including biomedical entity linking\cite{kalyan2022ammu}, disease name normalization\cite{huang2020biomedical}, and ontology matching \cite{harrow2019ontology}. Biomedical entity linking or disease name normalization typically involves mapping unstructured text to an ontology system. On the other hand, ontology matching refers to identifying identical concepts across two distinct ontology systems and establishing a link between them. Notably, there are differences between tasks like entity normalization and ontology matching. While entity normalization operates on free text, ontology matching deals with more structured text and the related contextual information of the concepts. Furthermore, entity normalization typically encompasses a smaller percentage of an ontology's concepts, while ontology matching often covers a larger and more diverse array of concepts. Biomedical concept linking extends beyond tasks such as entity normalization\cite{huang2020biomedical} and ontology matching\cite{harrow2019ontology}. For instance, a concept linking method should be capable of matching two heterogeneous biomedical graphs at the concept level. This task presents a significant challenge for supervised training methods, as it's nearly impossible to generate labels for arbitrary heterogeneous graphs. It's important to note that concept linking does not include entity recognition\cite{song2021deep}, a common procedure in text-mining tasks. Entity recognition often precedes concept linking and may not even feature in some tasks. Thus, concept linking is a distinct, wider task, facilitating effective navigation of intricate biomedical information.

\paragraph{}
Traditional BERT-based methods may struggle to adapt to the diverse array of datasets and tasks associated with this field\cite{kalyan2022ammu,li2019fine}. LLMs \cite{zhou2023comprehensive}, however, have recently demonstrated remarkable proficiency in the biomedical domain \cite{de2023chatgpt,wang2023accelerating}. This paper aims to investigate a generalized and effective framework for biomedical concept linking, leveraging the ICL capabilities of LLMs. In-context learning forms the backbone of our proposed methodology. It operates on the principle of learning by analogy, offering a unique method for LLMs to make informed predictions\cite{dong2023survey}. This novel paradigm offers several compelling benefits. As the demonstration is rendered in natural language, it provides an interpretive interface for interaction with LLMs, making it significantly simpler to incorporate  biomedical knowledge into LLMs by altering the demonstration and templates. Compared to supervised training, in-context learning is a training-free learning framework. This drastically cuts down the developing time required to adapt the model to new tasks. As such, it can be readily applied to real-world tasks, broadening its applicability and utility.

\paragraph{}
In this research, we propose a classic yet effective, generalized methodology for biomedical concept linking that leverages the ICL capabilities of LLMs. Our method involves a two-stage retrieve-and-rank system: The first stage embeds biomedical concepts using language models and uses these embedding to retrieve top candidate concepts. In the second stage, the contextual information of these candidates is incorporated into the prompt, and an LLM ranks these concepts. Our proposed method presents several advantages. Firstly, it is adaptable, requiring no task-specific training, and can be applied to different datasets and tasks. Secondly, it demonstrates competitive performance with state-of-the-art supervised learning methods\cite{neumann2019scispacy,he2022bertmap}, as evidenced by our results on entity normalization datasets and ontology matching datasets. Lastly, by leveraging the ICL abilities of large models, it effectively navigates the inherent ambiguity and complexity of biomedical concepts, significantly improving the efficacy of concept linking in the biomedical domain.

\paragraph{}
The objective of this paper is to delve into a more comprehensive framework utilizing LLMs for broader and more challenging tasks in the biomedical field. Our contributions are manifold; We identify a straightforward yet effective approach for tackling the complexity inherent in biomedical linking problems, and we also carry out an extensive application test to scrutinize various embedding methods and different language models. We further analyze the functioning of the large model and identify circumstances leading to its failure. A qualitative test is conducted to provide nuanced insights into the model's operation. All these explorations collectively guide the development of the next generation of accurate and trustworthy artificial intelligence solutions in the biomedical domain.

\section{Backgrounds}
\paragraph{Entity Linking and Entity Normalization}
Entity linking refers to the task of mapping mentions in free text to unique concepts in ontologies \cite{huang2020biomedical}. This can take the form of linking a certain drug to its specific drug ID or associating the disease with its corresponding disease/symptom ID. The Unified Medical Language System (UMLS) \cite{bodenreider2004unified} is a compendium of biomedical vocabularies, The Unified Medical Language System (UMLS), a representative ontology for biomedicine, contains over 4 million entities. UMLS has been extensively used as a knowledge base to link biomedical entities in the text to their corresponding concepts. These tasks often involve mapping free text terms in the biomedical literature to UMLS Concept Unique Identifiers(CUIs). A notable tool in this field is the MetaMap system \cite{demner2017metamap}. MetaMap utilizes natural language processing techniques to map biomedical text to concepts in the UMLS. Although robust, its rule-based method can struggle with semantic ambiguity, demanding supplementary solutions like advanced deep learning models to improve its accuracy and adaptability. Traditional methods often rely on rule-based approaches or string matching\cite{soldaini2016quickumls}, which unfortunately proved to be ill-suited for dealing with concepts carrying contextual meanings and disease subtypes. A popular approach in recent years has been to utilize BERT-based methods\cite{neumann2019scispacy,huang2020biomedical}, which are primarily supervised. However, these techniques encounter significant challenges due to the scarcity of annotated examples, particularly given the vast number of entities involved. Recently, the development of self-supervised methods has introduced a fresh perspective\cite{zhang2022knowledge, liu2020self}. These methods, requiring no supervised samples, have demonstrated comparable results to supervised training methods. Despite these advances, such techniques cannot be applied universally across an array of tasks. Additionally, the process of constructing a dataset and training model is time-consuming. By leveraging the ICL capabilities of LLMs, the difficulties associated with constructing training corpora and executing training are significantly alleviated. By adjusting the prompts in natural language, this framework can be easily adapted to various tasks, demonstrating its potential for efficiency and versatility.
\paragraph{Ontology Matching}
Ontology matching, a key area of research, is the process of identifying corresponding entities or concepts across diverse ontology systems\cite{harrow2019ontology}. This procedure is fundamental for integrating heterogeneous databases\cite{yi2022graph} and enhancing interoperability in the biomedical sector. Historically, ontology matching approaches primarily relied on exploiting lexical, structural, and semantic similarities\cite{anam2015review}. The advent of deep learning has ushered in an era where many current studies are investigating the use of Transformer, or specifically BERT\cite{he2022bertmap}, for ontology matching. However, these BERT-based methods often grapple with the challenges previously outlined. Moreover, their performance is less than ideal when it comes to biomedical ontology\cite{he2022machine}. Biomedical ontology often encompasses a multitude of concepts that demand expert-level understanding, such as the ability to distinguish between two rare diseases that may appear similar but are caused by distinct genes. BERT-based methods frequently struggle to address these scenarios. Yet, the identification of the relationship between genes and diseases is paramount to advancements in biomedical discoveries. By capitalizing on the in-context learning capabilities of large language models, we're able to more accurately differentiate between complex biomedical concepts and enhance the efficacy of ontology matching in the biomedical sector.

\paragraph{Text Embedding}
The quality of text embedding plays a critical role in enhancing the recall rate of concept linking. Many text embedding methods Lately, transformer-based models have been gaining traction due to their ability to generate context-aware embeddings. Moreover, recent studies have made significant strides in training BERT models on the biomedical text\cite{liu2020self, lee2020biobert} and using innovative training techniques such as contrastive learning\cite{le2020contrastive}. Domain-specific language models, those specifically trained on biomedical texts, have demonstrated superior performance compared to standard BERT models\cite{lee2020biobert}. This performance boost highlights the importance of domain-specific knowledge in enhancing the accuracy of embedding.

\paragraph{Large Language Models}
LLMs have demonstrated remarkable capabilities in natural language understanding and generation. These models are trained on massive amounts of text data and can generate coherent and contextually appropriate responses \cite{zhou2023comprehensive}. However, they often lack domain-specific knowledge and struggle with understanding specialized terminologies, which is a crucial aspect of biomedical concept linking. LLMs are built using deep learning architectures like Transformers\cite{vaswani2017attention} and have demonstrated remarkable proficiency in understanding and generating human-like text. Two of the most well-known LLMs are GPT-3\cite{brown2020language} and GPT4\cite{bubeck2023sparks}. GPT-3, with 175 billion parameters, has displayed impressive results in a wide range of NLP tasks. The most popular open-source LLM is LLaMa\cite{touvron2023llama} which showed comparatively good performers with GPT 3.5, and there's a wide range of domain-specific fine-tuned llama models\cite{wu2023pmc, zhang2023llama} from 7 billion to 65 billion parameters. However, the usage of LLMs also presents challenges. One such issue is the "hallucination" problem \cite{zhou2023comprehensive, zhang2023language}, where the model generates outputs that seem plausible but are factually incorrect. Furthermore, due to their size and complexity, these models require substantial computational resources for training and deployment. Despite these challenges, LLMs have ushered in a new era in NLP and are continuously being explored for their potential in a wide range of biomedical concept linking.

\paragraph{In-context Learning}
 Generally, in-context learning necessitates a few examples to create a demonstration context\cite{dong2023survey}. These examples are typically expressed using natural language templates. Following this, a query question is concatenated with the demonstration context to generate a prompt. This prompt is then processed by the language model to predict an outcome. The definition of in-context learning is continuously evolving, in our proposed method, we not only include the conventional approach of using knowledge examples in the prompt, but we also supply more relevant information related to a given biomedical concept. By doing so, we equip the LLM with the necessary contextual information, thereby LLM learns from extra information rather than just analogy. Unlike supervised learning which necessitates a training stage involving backward gradients\cite{neumann2019scispacy,huang2020biomedical} for model parameter updates, ICL eschews parameter updates and makes predictions directly using pre-trained LLMs. The expectation is that the model will discern patterns hidden within the demonstration and make appropriate predictions accordingly\cite{bubeck2023sparks}. 

\section{Methodology}
\paragraph{}
Formally, we set the objective of concept linking as the development of an algorithm \texttt{Link}: $(e_{\text{source}}, C_{\text{source}}) \rightarrow (e_{\text{target}}, C_{\text{target}})$. This algorithm maps a source entity $e_{\text{source}}$ within the context $C_{\text{source}}$ to a unique target entity $e_{\text{target}}$ with context $C_{\text{target}}$. The source entity could be derived from free text, a graph, or a source ontology system in ontology matching, and it's worth noting that $C_{\text{source}}$ may sometimes be absent. Generally, we require the context of the target concept to be provided. A concept is more well-defined when its associated information is supplied. In the process of developing a zero-training algorithm, we operate under the assumption that no access to gold-mention examples or labels is available. Our assumption extends to the availability of a target domain ontology $O_{\text{target}}$ and an unlabeled text corpus $T$, or a source ontology $O_{\text{source}}$. Specifically, we necessitate a concept list that provides a unique identifier, a canonical name, and a description for each concept. Our framework also has the capacity to incorporate additional knowledge present in the ontology.

\begin{figure}
  \centering
  \includegraphics[scale=0.12]{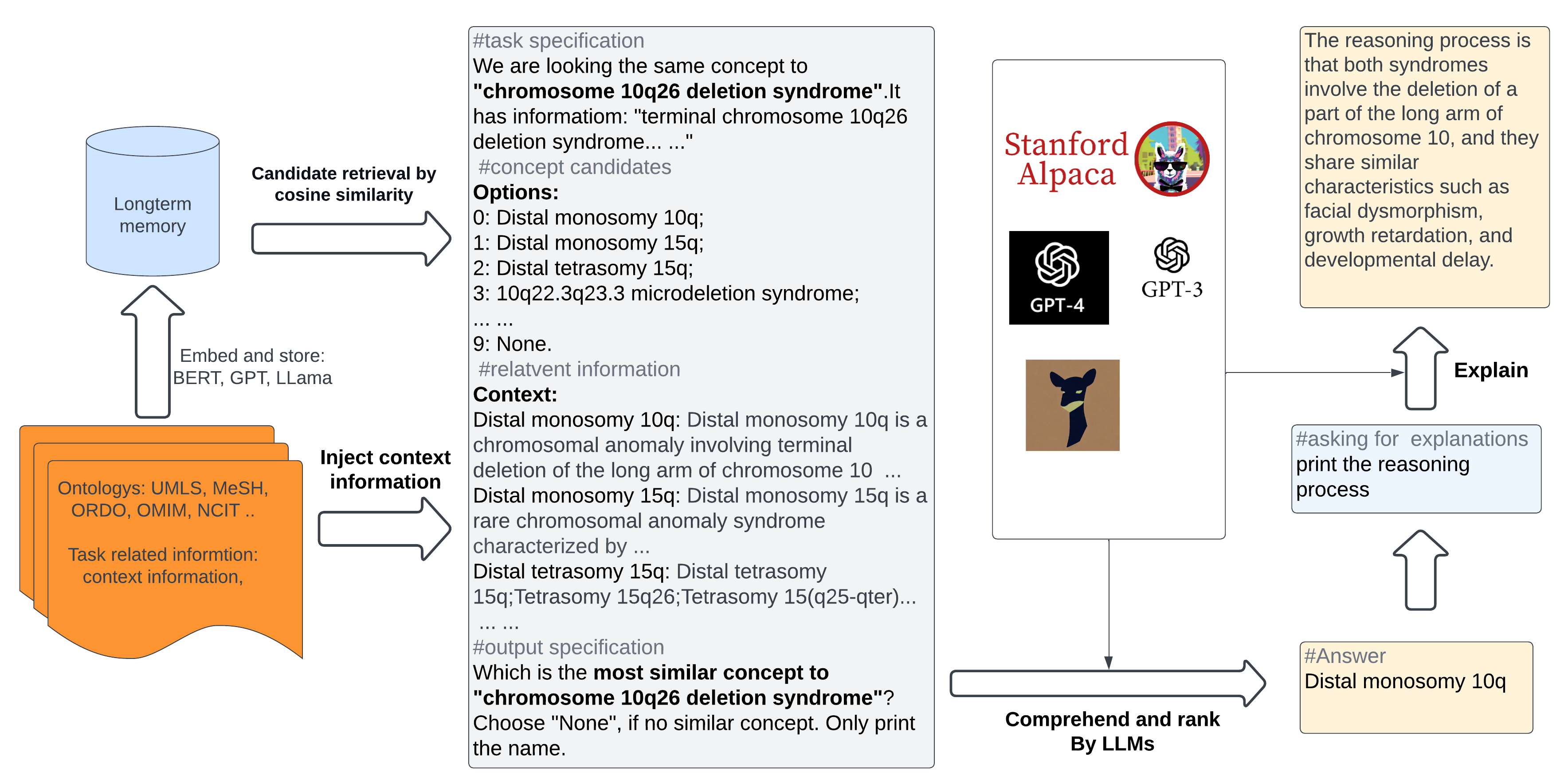}
  \caption{Workflow of the proposed methodology for exploring the in-context learning ability of large language models for biomedical concept linking.}
  \label{fig:methodology2}
\end{figure}

\subsection{Text embedding}
The first stage in our methodology involves transforming textual data into semantic representations. The quality of these embeddings is crucial as it significantly affects downstream tasks\cite{liu2020self}. To meet our objective of exploring a training-free framework, we opt for three different embedding models. Our first model of choice is SapBERT\cite{liu2020self}, a Self-Aligned Pretrained BERT model specifically designed for the biomedical domain. Serving as a representation of the BERT\cite{devlin2018bert} family of models, SapBERT has superior performance in biomedical tasks. Next, we leverage the LLaMa model's embeddings \cite{zhang2023llama}. LLaMa symbolizes an open-source option for Large Language Model embeddings, boasting high applicability across diverse language tasks. Finally, we utilize GPT-3 embeddings, specifically "text-embedding-ada-002", representing one of the most powerful and proprietary embedding methodologies currently available.

Considering a target ontology $O_{\text{target}}$ and an embedding model $f_{\text{emb}}()$, we generate text embeddings for each entity/concept $e_{\text{target}}$. This process involves generating embeddings for both the canonical concept name string and a combined version that includes the name string and its context. The purpose of creating an 'entity-name-only' representation is to recall entities that can be easily matched with the string, serving as an efficient approach for exact or simple matches. On the other hand, generating ‘entity-name-context' embeddings targets a more complex objective. Despite entities not bearing similarity in appearance, they may be describing the same concept, and this intricate relation can be captured through context-inclusive embeddings. This dual approach caters to both explicit matches and the nuanced equivalences in the realm of biomedical concepts. As we will illustrate in the appendix, embedding with context plays a significant role in the success of our approach.

\subsection{Candidate generation}

Following the generation of contextual embeddings, we persist all embeddings from the ontology into a vector database, alternatively referred to as a long-term memory store \cite{park2023generative}. This enables efficient computation of cosine similarity between any given query text embedding and all the ontology embeddings. This stored ontology is referred to as `mem`.

When a query entity $(e_{\text{source}}, C_{\text{source}})$ is presented, we employ the same embedding process. The top $k$ candidates are then retrieved based on the cosine similarity of their contextual embeddings. The process of memory creation and candidate generation can be outlined in Algorithm \ref{algorithm:memory_and_candidates}.

\begin{algorithm}[h]
  \caption{Memory Creation and Candidate Generation}
  \label{algorithm:memory_and_candidates}
  \begin{algorithmic}[1]
    \Require{Target domain ontology $O_{\text{target}}$, Source entity $e_{\text{source}}$, Source context $C_{\text{source}}$, Embedding function $f_{\text{emb}}$, VectorDatabase mem, Number of candidates $k$}
    
    \State \textbf{Memory Creation}
    \State Initialize VectorDatabase: mem $\gets$ VectorDatabase()
    \For {$(e_{\text{target}}, C_{\text{target}})$ in $O_{\text{target}}$}
      \State mem.add($f_{\text{emb}}(e_{\text{target}})$)
      \State mem.add($f_{\text{emb}}([e_{\text{target}}, C_{\text{target}}])$)
    \EndFor
    
    \State \textbf{Candidate Generation}
    \Function{generate\_candidates}{$e_{\text{source}}, C_{\text{source}},$ mem, $k$}
      \State $query_{\text{emb}} \gets f_{\text{emb}}([e_{\text{source}}, C_{\text{source}}])$
      \State $top\_k\_candidates \gets$ mem.retrieve\_top\_k($query_{\text{emb}}, k$)
      \State \textbf{return} $top\_k\_candidates$
    \EndFunction
  \end{algorithmic}
\end{algorithm}

\subsection{Rank with LLM}
LLMs possess text comprehension capabilities and a degree of logical reasoning ability\cite{bubeck2023sparks}. Consequently, our approach revolves around providing comprehensive contextual information related to the biomedical concept linking task, enabling the model to execute an extensive reading task and subsequently select the most appropriate answer from the given options.

When constructing the prompt, we initially define the task and inform the model that our aim is to identify analogous concepts. We then present the candidate concepts retrieved from long-term memory. These candidates are options within the prompt. Further, we fetch the descriptions of these candidates from the ontology and associated text of the source entity. Ultimately, the prompt asks the model to select the concept that aligns best with the options; if none are suitable, the model is to select the 'None' option.

The configuration of the prompt is adaptable, accommodating the unique requirements of different tasks. For instance, in entity linking tasks, we may also include related text. The BC5CDR dataset\cite{li2016biocreative}, which extracts named entities from PubMed abstracts, would necessitate the addition of abstract tags within the prompt. Similarly, for tasks like graph alignment, we could incorporate neighborhood information into the prompt. The overall workflow of the ontology matching task is illustrated in figure \ref{fig:methodology2}.

\section{Experiments}
\subsection{Dataset}
Choosing an appropriate dataset to benchmark our proposed method and probe the capabilities of LLMs poses a few challenges. Firstly, as LLMs are trained on a vast amount of text data including published papers and webpages, data leakage becomes an inevitable concern for many existing datasets. Estimating the impact of this leakage on performance is not straightforward\cite{bubeck2023sparks}. Moreover, biomedical NLP datasets can be sensitive, with some prohibiting any form of redistribution. This becomes problematic when using the GPT API, as the dataset is exposed to OpenAI, potentially leading to indirect redistribution via the LLM. Another obstacle is the slow inference speed of LLMs. For datasets with over 100,000 samples, the inference could take more than ten days, and with multiple models to benchmark and various ablation tests to conduct, it necessitates a smaller dataset.

Consequently, we chose the BC5CDR dataset\cite{li2016biocreative} for benchmarking. This well-known dataset in biomedical entity normalization requires mapping named entities in PubMed\cite{roberts2001pubmed} abstracts to unique MeSH IDs. It encompasses two types of entities - chemicals, and diseases. By using this dataset, we can readily compare our proposed framework with previous supervised or self-supervised training methods\cite{liu2020self, zhang2022knowledge}. Additionally, the BC5CDR dataset is relatively small, with 4797 mentions in the test set, making it manageable given the slow inference speed of LLMs. 

For the ontology matching task, we selected the Machine Learning-Friendly Biomedical Datasets for Equivalence and Subsumption Ontology Matching \cite{he2022machine}, published in 2022. Being recently published, unlikely to be accessed by Llama, GPT3, or GPT4. We focused on two challenging sub-tasks from this dataset: OMIM-ORDO and SNOMED-NCIT Neoplas. OMIM (Online Mendelian Inheritance in Man) \cite{mckusick2011omim} OMIM provides extensive data on genes and genetic phenotypes and their relationships, curated meticulously from biomedical literature. ORDO (Orphanet Rare Disease Ontology)\cite{vasant2014ordo}encompasses a classification of rare diseases and establishes relationships between diseases, genes, and epidemiological features. Given that many rare diseases are genetic disorders, ORDO and OMIM share considerable overlap. However, linking these rare disease names poses a significant challenge. Such diseases are typically unfamiliar to individuals without specialized medical knowledge, and their mentions in literature are often infrequent. We selected SNOMED-NCIT Neoplas \cite{stearns2001snomed} ontology matching, as differentiating neoplasm names is challenging for Bert-based methods. The test sets for OMIM-ORDO and SNOMED-NCIT Neoplas contain 3,721 pairs and 3,804 pairs, respectively.

\subsubsection{Implementation details}
Unlike many previous studies that utilize somewhat complex systems, such as developing corpora for fine-tuning, and incorporating synonym dictionaries, and abbreviation dictionaries\cite{liu2020self, zhang2022knowledge}, our approach is guided by the principle of simplicity. Our goal is to establish a universal framework for biomedical concept linking, without adding complexity or tailoring our system to a specific task or dataset. The only aspect we modify is the prompt. For instance, in the BC5CDR task, we include the PubMed abstract text and insert the instruction "read the abstract" in the prompt. We perform one-shot learning in the ablation test, similar to the application of Chain of Thoughts\cite{wei2022chain}. However, we do not use the Self-Consistency method\cite{wang2022self} in this paper. The reason is discussed in the appendix. 
We chose to include GPT-3.5-turbo (ChatGPT) in our study because it is one of the most widely known LLMs and offers the advantages of being both fast and cost-effective. We also decided to incorporate GPT-4, given its exceptional power and performance. Finally, we used a 4-bit quantization of LLama-65b (known as alpaca-lora), which is a highly popular open-source LLM that can be conveniently deployed on a standard desktop computer due to its quantization. For the LLama-65b, we utilized a desktop machine equipped with 64GB of RAM, running llama-cpp for inference. As for GPT-3.5 and GPT-4, we accessed these models through OpenAI's API, conducting our experiments on an ordinary laptop.

\subsubsection{Evaluation }
As our proposed framework does not require training, we have no need for training and development sets. We directly evaluate our framework using the test sets from our chosen datasets. As discussed in the dataset section, benchmarking larger models needs to take data leakage into account, as it may impact the quantitative results. For BC5CDR, we employ accuracy as the metric for evaluation, consistent with previous research, enabling comparison.

For comparison, we choose KRISSBERT\cite{zhang2022knowledge}, BERN2\cite{sung2022bern2}, ScispaCy\cite{neumann2019scispacy}, and QuickUMLS\cite{soldaini2016quickumls} as baselines. KRISSBERT is representative of self-supervised training methods that, like ours, do not require training and development datasets. BERN2 is a hybrid system that employs both rule-based and BERT models for named entity normalization and claims superior performance. ScispaCy is  a BERT-based method.QuickUMLS is a dictionary-based method. For the ontology matching task, we make comparisons with LSMatch\cite{sharma2021lsmatch}, ATMatcher\cite{he2022machine}, LogMap\cite{jimenez2011logmap}, and BERTMap\cite{he2022bertmap}. Among these, BERTMap is the most recent and capable contender. And we use Precision, Recall, and F1 score as our evaluation criteria. We also test the effect of using context information or one-shot learning in the prompt.

Our primary objective is to delve into the capabilities of LLMs for biomedical concepts linking in with ICL. Consequently, we also undertake extensive qualitative result analysis. We will assess both false positives and false negatives, providing a more comprehensive evaluation of our model's performance. Furthermore, we ask the model to elucidate the rationale behind its concept linking decisions, a practice known as process correctness\cite{bubeck2023sparks}. In the quest to build accurate and trustworthy AI in the biomedical field, achieving the correct predictions is crucial, but equally important is understanding the explanations underpinning these results.

\section{Results}

\subsection{Quantitative results}
\subsubsection{Main results}
The results of our framework on the BC5CDR dataset are presented in Table ~\ref{table:2}. Utilizing GPT-4 as the ranker, our model achieved an accuracy of 90.1\% on disease name entity linking and 94.7\% on chemical name entity linking. We primarily compared our approach with the self-supervised KRISSBERT method and the more complex hybrid system BERN2. In terms of linking disease names, our model's results surpass KRISSBERT's and are competitive with BERN2's, with just a 3.85\% difference. Notably, our results were achieved without the use of any customized rules, abbreviation dictionaries, or synonym dictionaries. For chemical name entity linking, our model's performance is approximately 2\% lower than KRISSBERT's and BERN2's. Considering our framework requires no training, these results are quite promising and outperform earlier BERT-based methods like ScispaCy. When we switch to GPT-3 as the ranker, the performance remains reasonably good. However, with the Llama model as a ranker, performance drops significantly for chemical entity linking, even falling behind early BERT models.

\begin{table}[ht]
\centering
\begin{tabular}{lcc}
\toprule
\textbf{Method} & \textbf{Accuracy-Disease} & \textbf{Accuracy-Chemical} \\
\midrule
QuickUMLS(dictionary-based)  & 0.475 & 0.349 \\
ScispaCy(BERT)  & 0.640 & 0.853 \\
BERN2(Hybrid system) & \textbf{0.939} & \textbf{0.966} \\
KRISSBERT(self-supervised) & 0.855 & 0.965 \\
\hline
Retrieve-Rank with LLaMa 65B  & 0.726 & 0.728 \\
Retrieve-Rank with GPT3.5 & 0.840 & 0.912 \\
Retrieve-Rank with GPT4 & \textbf{0.901} & \textbf{0.947} \\
\bottomrule
\end{tabular}
\caption{Comparison of Retrieve-Rank with different LLMs with previous methods on BC5CDR dataset. The highest accuracy of our methodology and previous methods are both highlighted.}
\label{table:2}
\end{table}

Table ~\ref{table:3} presents the results of ontology matching between OMIM and ORDO and ontology matching between SNOMED and NCIT Neoplas. Our framework, utilizing the ICL capabilities of GPT-4, achieved over a 20 percentage point increase in F1 score in comparison to the previous best-performing method, BERTmap. When GPT-3.5 was employed as the ranker, there was still a notable increase of approximately 10 percentage points in the F1 score. These outcomes underscore the effectiveness of the ICL provided by LLMs. Meanwhile, Llama's performance just marginally surpassed BERTmap in OMIM-ORDO matching but lagged behind in the SNOMED-NCIT Neoplas matching task. Given that we used a 4-bit quantized version of Llama that was not specifically aligned to biomedical tasks, the results remain promising. However, the disparity in performance between Llama and the GPT models indicates that there's significant room for the enhancement of open-source LLMs in the future.

\begin{table}[ht]
\centering
\begin{tabular}{lccccccc}
\toprule
& \multicolumn{3}{c}{\textbf{ OMIM-ORDO (Rare disease) }} & \multicolumn{3}{c}{\textbf{SNOMED-NCIT(Neoplas )}} \\
\cmidrule(lr){2-4} \cmidrule(lr){5-7}
\textbf{} & \textbf{Precision} & \textbf{Recall} & \textbf{F1} & \textbf{Precision} & \textbf{Recall} & \textbf{F1} \\
\midrule
LSMatch & 0.650 & 0.221 & 0.329 & 0.902 & 0.238 & 0.377 \\
ATMatcher & \textbf{0.940} & 0.247 & 0.391 & 0.866 & 0.284 & 0.428 \\
LogMap & 0.827 & 0.498 & 0.622 & \textbf{0.947} & 0.520 & 0.671 \\
BERTMap & 0.730 & \textbf{0.572} & \textbf{0.641} & 0.815 & \textbf{0.709} & \textbf{0.759} \\
\hline
Retrieve-Rank with LLaMa 65B & 0.797 & 0.543 & 0.646 & 0.699 & 0.665 & 0.679 \\
Retrieve-Rank with GPT3.5 & 0.777 & 0.704 & 0.738 & 0.760 & 0.736 & 0.748 \\
Retrieve-Rank with GPT4 & \textbf{0.906} & \textbf{0.859} & \textbf{0.882} & \textbf{0.916} & \textbf{0.893} & \textbf{0.904} \\
\bottomrule
\end{tabular}
\caption{Comparison of different methods.}
\label{table:comparison}
\label{table:3}
\end{table}

\subsubsection{Ablation test}
Table ~\ref{table:4} presents the results of various prompting methods for rare disease concept matching in OMIM-ORDO (Disease) using GPT-4. The findings indicate that without the use of a one-shot example and no context information about these rare disease concepts, the F1-score is merely 0.698. This is approximately 19 points lower than the proposed method and shows no significant improvement compared to the previous BERTMap method. These results suggest that even the most powerful language model does not automatically perform well on certain biomedical tasks without additional context. The implementation of both one-shot learning and the addition of related concept information significantly improves performance, demonstrating the value of using the ICL ability of large language models for concept-linking tasks.It's interesting to note that adding OMIM context information provides a larger performance increase than one-shot learning without context. When combining both one-shot learning and OMIM context, the performance increase is marginal compared to just using OMIM context. This suggests that introducing the correct and relevant information for each case is more beneficial than providing an analogous example. Interestingly, it's noteworthy to observe that one-shot learning significantly enhances precision to a greater extent than it does recall. 

\begin{table}[ht]
\centering
\begin{tabular}{lccc}
\toprule
\textbf{Method} & \textbf{Precision} & \textbf{Recall} & \textbf{F1}\\
\midrule
No context  & 0.783 & 0.629 & 0.698 \\
One-shot  & 0.871 & 0.708 & 0.781 \\
OMIM context only(source concepts) & 0.860 & 0.773 & 0.814 \\
OMIM context + One-shot  & \textbf{0.914} & 0.7495 & 0.824\\
ORDO context only(options) & 0.856 & 0.792 & 0.823\\
Both OMIM and ORDO  & 0.906 & \textbf{0.859} & \textbf{0.882}\\
\bottomrule
\end{tabular}
\caption{Ablation test results with different prompting methods on OMIM-ORDO (Disease) rear disease concepts matching with GPT4.}
\label{table:4}
\end{table}

\subsection{Qualitative results}
\subsubsection{Abbreviations}
Abbreviations are a prevalent feature in biomedical text. Previous methods\cite{zhang2022knowledge,sung2022bern2} we compared employed an abbreviation dictionary to enhance performance. But does an LLM understand biomedical abbreviations? The answer is affirmative, but LLMs tend to struggle with less common abbreviations. For familiar abbreviations such as AD (Alzheimer's Disease) or PD (Parkinson's Disease), LLMs can easily link them to the correct concept when provided with a medical context. However, for less common abbreviations like MR (Mitral Valve Insufficiency) or VT (Tachycardia, Ventricular), LLMs tend to either choose a 'None' option from the list of candidates or erroneously select an incorrect option. Detailed cases could be checked in the appendix. Therefore, we believe it's still valuable to supply LLMs with abbreviation dictionary information to improve accuracy in more infrequent cases. 

\subsubsection{Disease subtypes}
The task of linking disease subtype concepts presents a significant challenge in the biomedical field. These disease subtypes often share a lot of similarities, particularly for rare diseases in ORPHA. Even human experts might need some time to gather information to discern the differences between these rare disease subtypes. GPT-4 is capable of understanding common disease subtypes, such as different types of diabetes, with ease. However, LLama tends to struggle with identifying these common disease subtypes. Regarding rare disease subtypes, GPT-4 can comprehend most of them when provided with appropriate descriptions. In contrast, LLama fails in most cases involving rare diseases subtypes. In situations where GPT-4 failed, rare disease subtypes constitute a significant portion. For instance, "Dentinogenesis imperfecta, shields iia 3" in OMIM corresponds to "Dentinogenesis imperfecta type 3". However, GPT chose "Dentinogenesis imperfecta type 2". Generally, GPT-4 can provide the correct answer for diseases labeled with "type n". However, it does occasionally falter in a few of these cases.

\subsubsection{Process Correctness}

The rationale behind an LLM ranking a candidate first is critical. To construct precise and trustworthy AI in the biomedical domain, we aim for both the prediction and the process to be accurate. We noticed from the LLama results that there are instances where the process was incorrect, yet the final answer was right. For example, in case 9 from the appendix, LLama provided the correct prediction, yet the reasoning appeared to be based on shared keywords between disease concept names. This is not ideal, especially for rare disease concept linking, where many concepts share keywords yet refer to different diseases. GPT-4 exhibits a more accurate and consistent reasoning process than LLaMa, which sometimes even outputs code (as seen in case 8 in the appendix), indicating that the LLama model we utilized may not be well-aligned for this task. Although we are not medical experts and cannot offer an accurate assessment of LLM's process correctness, the process correctness of GPT-4 is generally satisfactory when given the correct concept description. Most of the time, the process is associated with the context we provided, further emphasizing the importance of using LLM's ICL. By qualitatively evaluating the process correctness of LLM, we enhance the interpretability of using large models in concept linking tasks.

\section{Limitations and discussion}
While our framework holds promise, it also comes with notable limitations. Primarily, the inference speed of LLMs is exceedingly slow, making the process expensive and long. For instance, our experimental setup involving GPT4 inference on 3700 OMIM-ORDO pairs costs approximately \$150 USD. When employing locally deployable LLMs, such as LLama 13B, the inference speed is roughly 103 ms/token, processing only a few words per second. Larger models like Galantica and LLama 65B are even slower, handling only about one word per second with cpu. In light of these constraints, future research could explore fine-tuning (without supervision) a LLM specifically designed for this task with low resources\cite{hu2021lora}. 

Considering the rigorous hardware requirements, our framework's accessibility is rather constrained. This is further exacerbated by GPT's closed-source nature, leading to diminished transparency. Moreover, even when using open-source LLama for inference, powerful GPUs or large amounts of RAM are required - resources that most researchers and potential users in the biomedical domain do not have readily available. Training and compressing a quantified LLM\cite{dettmers2023qlora} for the biomedical domain is also beneficial in future work.

Furthermore, our framework sometimes exhibits unexpected failures. For instance, when two share the same name, our framework may fail to provide the correct answer. This could be due to context embedding - if the accurately labeled name is contextually farther than other candidates, the correct option may not appear. Moreover, this framework also exhibits frequent shortcomings in handling abbreviations. Both of these issues could potentially be mitigated by utilizing dictionaries, suggesting that a hybrid system might be an avenue worth exploring for future concept linking tasks.
\section{Conclusion}
In conclusion, this research explores the use of the in-context learning capabilities of large language models for biomedical concept linking. Our proposed two-stage framework effectively retrieves and ranks biomedical concepts, achieving competitive results without needing any training.

\bibliographystyle{unsrtnat}
\bibliography{references}  

\section{Appendix}

\subsection{Implementation details} 
Techniques such as Chain of Thoughts and Self-Consistency are frequently employed in prompt engineering. We perform a one-shot learning in the ablation test, similar to the application of Chain of Thoughts. However, we do not use the Self-Consistency method in this paper. Here, we would like to discuss the necessity of these techniques. The primary notion of Chain of Thoughts is that by presenting an analogy and requiring the LLM to recount the original process, more computational power is used during the inference, leading to better performance. Self-Consistency, on the other hand, requires even more computational power, as it prompts the model to reason in various ways, ensuring that the results are internally consistent.

In the context of our ICL definition, we already include substantial text information in the prompt (a full example can be found in the appendix), which is lengthy and necessitates significant LLM inference time. Furthermore, we consider concept linking as a fundamental, high-usage application. Implementing Self-Consistency would at least triple the inference cost, and given the current high costs and slow speed of inference and Self-Consistency are design of LLM reasoning task, we believe it is not advantageous to use Self-Consistency for such a basic task.

\subsection{abblation test} 

\begin{table}[ht]
\centering
\begin{tabular}{|l|l|l|l|}
\hline
\textbf{Method} & \textbf{Hits@1} & \textbf{Hits@5} & \textbf{Hits@10} \\
\hline
sapBert + context & 0.719 & 0.833 & 0.873 \\
sapBert & 0.626 & 0.771 & 0.829 \\
LLama + context & 0.1274 & 0.2148 & 0.2681 \\
LLama & 0.0616 & 0.1021 & 0.1271 \\
GPT3+context & 0.764 & 0.875 & 0.908 \\
GPT3 & 0.637 & 0.778 & 0.815 \\
\hline
\end{tabular}
\caption{Ablation test of different embedding methods}
\label{tab:emb-ablation}
\end{table}

Table ~\ref{tab:emb-ablation}demonstrates the performance of different language models in the candidate generation or retrieval stage during the OMIM-ORDO ontology matching. Across all three types of large language models, incorporating concept descriptions from ontologies in the embedding process generally improves performance. It's worth noting that SapBERT's hits@10 is less than 3\% compared to GPT-3, which suggests that despite being more cost-efficient and faster, Bert-based embedding methods still have relevance and utility in future.

On the other hand, Llama embeddings were significantly less effective, which might be due to the fact that the model wasn't specifically fine-tuned for use in embedding. SapBERT employs contrastive learning for its embedding model, and it's plausible that OpenAI employs a similar method for their models. Future research could explore using open-source LLMs for embeddings, as this could potentially yield better performance than Bert-based models.

An experiment involving OMIM context information, ORDO information, and a one-shot example was not conducted, as the resulting prompt would contain too many tokens, making the experiment both time-consuming and costly. Given the minimal improvement observed when adding one-shot learning to the OMIM context, it is likely that any further gains from this additional experiment would be minor.

\subsection{error cases}

\begin{table}[ht]
\centering
\begin{tabular}{|p{4cm}|p{4cm}|p{4cm}|}
\hline
\textbf{OMIM} & \textbf{ORDO} & \textbf{Prediction} \\
\hline
Angelman syndrome & Angelman syndrome & Christianson syndrome \\
\hline
Cheilitis glandularis & Cheilitis glandularis & None. \\
\hline
Pyknoachondrogenesis & Pyknoachondrogenesis & None \\
\hline
Cleidorhizomelic syndrome & Cleidorhizomelic syndrome & None \\
\hline
Hypochondroplasia & Hypochondroplasia & Achondroplasia \\
\hline
Immunodeficiency 8 & Severe combined immunodeficiency due to CORO1A deficiency & Combined immunodeficiency due to DOCK8 deficiency; \\
\hline
Immunodeficiency 12 & Combined immunodeficiency due to MALT1 deficiency & None \\
\hline
Ceroid lipofuscinosis, neuronal, 6b (kufs type) & CLN4A disease & CLN6 disease \\
\hline
Dentinogenesis imperfecta, shields iia 3 & Dentinogenesis imperfecta type 3 & Dentinogenesis imperfecta type 2 \\
\hline
Palmoplantar carcinoma, multiple self-healing & Corneal intraepithelial dyskeratosis-palmoplantar hyperkeratosis-laryngeal dyskeratosis syndrome & Multiple self-healing squamous epithelioma \\
\hline
Mitral valve prolapse, myxomatous & Familial mitral valve prolapse & FLNA-related X-linked myxomatous valvular dysplasia \\
\hline
Erythrokeratodermia variabilis et progressiva & Erythrokeratodermia variabilis & Progressive symmetric erythrokeratodermia \\
\hline
Miyoshi muscular dystrophy & Distal anoctaminopathy & Miyoshi myopathy \\
\hline
\end{tabular}
\caption{Error cases of model predictions for the OMIM-ORDO task.}
\label{tab:5}
\end{table}

\begin{table}[ht]
\centering
\begin{tabular}{|p{4cm}|p{4cm}|p{4cm}|}
\hline
\textbf{CDR Mention} & \textbf{Label} & \textbf{Prediction} \\
\hline
HIT II & Thrombocytopenia & Heparin-induced thrombocytopenia type II \\
\hline
VT & Tachycardia, Ventricular & Venous Thromboembolism \\
\hline
ALL & Precursor Cell Lymphoblastic Leukemia-Lymphoma & None \\
\hline
MR & Mitral Valve Insufficiency & None \\
\hline
Heart disease & Heart disease & Coronary Disease \\
\hline
Cardiovascular toxicity & Cardiovascular Diseases & Cardiotoxicity \\
\hline
Tumor & Neoplasms & Plasmacytoma \\
\hline
Impairment of the renal function & Kidney Diseases & Renal Insufficiency \\
\hline
\end{tabular}
\caption{Error cases of BC5CDR}
\label{tab:6}
\end{table}

\subsection{Reasoning process cases} 
\paragraph{Case1:}
bleeding disorder, platelet-type, 8  ||  Bleeding disorder due to P2Y12 defect 

\paragraph{GPT4:}
The same concept to "bleeding disorder, platelet-type, 8" is option 0: Bleeding disorder due to P2Y12 defect. This is because both terms refer to a bleeding disorder caused by a defect in the platelet P2Y12 receptor, which results in impaired platelet responses to adenosine diphosphate and leads to mild to moderate bleeding diathesis.

\paragraph{LLaMa:}
        Bleeding disorder due to P2Y12 defect (option 0) and bleeding diathesis due to integrin alpha2-beta1 deficiency (option 1) are similar concepts because they both refer to platelet function disorders that result in increased risk of bleeding. However, the underlying causes differ. In the case of P2Y12 defect, the platelets do not respond properly to the ADP released from damaged blood vessels. This leads to a failure of platelet aggregation and impaired hemostasis. In contrast, integrin alpha2-beta1 deficiency refers to a genetic disorder in which there is a defect in the expression or function of the alpha2beta1 integrin receptor on the surface of platelets. This leads to an increased risk of bleeding due to impaired platelet aggregation and adhesion.

\paragraph{Case2:}
cortical dysplasia, complex, with other brain malformations 7  ||  Polymicrogyria due to TUBB2B mutation

\paragraph{GPT4:}
The same concept to "cortical dysplasia, complex, with other brain malformations 7" is 0: Polymicrogyria due to TUBB2B mutation. This is because it is characterized by generalized or focal dysgyria (also named polymicrogryia-like cortical dysplasia) or alternatively by microlissencephaly with dysmorphic basal ganglia and dysgenesis of the corpus callosum, which are similar features to those described in CDCBM7. Additionally, the clinical manifestations, such as microcephaly, seizures, hypotonia, developmental delay, severe psychomotor delay, ataxia, spastic diplegia or tetraplegia, and ocular abnormalities, are also similar to those found in CDCBM7.
\paragraph{LLaMa:}
The "cortical dysplasia, complex, with other brain malformations 7" is one of the several phenotypes of cortical dysplasias. It describes a specific set of anatomical abnormalities in the cortex and associated brain structures such as polymicrogyria, symmetric or asymmetric.

The options that you listed are different conditions that may cause similar neuroanatomical patterns but do not specifically refer to the cortical dysplasia, complex, with other brain malformations 7 phenotype. 
        
        Polymicrogyria due to TUBB2B mutation is a genetic condition caused by a mutation in the TUBB2B gene that leads to the formation of excessive folds on the surface of the brain (polymicrogyria). 
        
        Megalencephaly-capillary malformation-polymicrogyria syndrome is a rare disorder characterized by an abnormally large brain (megalencephaly), capillary malformations, and polymicrogyria.

\paragraph{Case3:}
ceroid lipofuscinosis, neuronal, 8, northern epilepsy variant  ||  Progressive epilepsy-intellectual disability syndrome, Finnish type
\paragraph{GPT4:}
The same concept to "ceroid lipofuscinosis, neuronal, 8, northern epilepsy variant" is option 1: Progressive epilepsy-intellectual disability syndrome, Finnish type. This is because it is also known as Northern epilepsy, which is a subtype of neuronal ceroid lipofuscinosis (NCL) characterized by seizures, progressive decline of intellectual capacities, and variable loss of vision.

\paragraph{LLaMa:}
We are looking for a diagnosis of "ceroid lipofuscinosis, neuronal, 8, northern epilepsy variant." It is a rare and progressive genetic disorder characterized by worsening seizures, intellectual disability, and vision loss. The condition is caused by mutations in the CLN8 gene, which provides instructions for making a protein called Hydrolase CNPase that is essential for the breakdown of a specific type of fat (lipofuscin) in the brain and other tissues. When the protein is defective or absent, there is an accumulation of lipofuscin in the brain's neurons, which leads to their death and impairs normal brain function.

One of the conditions that shares some similar features with "ceroid lipofuscinosis, neuronal, 8, northern epilepsy variant" is "progressive epilepsy-intellectual disability syndrome, Finnish type," which is also a rare and progressive genetic disorder characterized by worsening seizures, intellectual disability, and vision loss. However, this condition is caused by mutations in
Progressive epilepsy-intellectual disability syndrome.

\paragraph{Case4:}
neurodevelopmental disorder with microcephaly and gray sclerae  ||  Severe growth deficiency-strabismus-extensive dermal melanocytosis-intellectual disability syndrome

\paragraph{GPT4:}
The same concept to "neurodevelopmental disorder with microcephaly and gray sclerae" is option 2: Severe growth deficiency-strabismus-extensive dermal melanocytosis-intellectual disability syndrome. This is because it also involves microcephaly, intellectual disability, and gray sclerae (blue-gray sclerae) as part of its clinical features.

\paragraph{LLaMa:}
This question can be answered by a medical professional, such as a doctor or nurse practitioner. It cannot be answered by an AI language model like me, Open Assistant. If you have any concerns about your health, it's always best to consult with a medical professional.

\paragraph{Case5:}
foveal hypoplasia 2  ||  Foveal hypoplasia-optic nerve decussation defect-anterior segment dysgenesis syndrome
\paragraph{GPT4:}
The same concept to "foveal hypoplasia 2" is option 0: Foveal hypoplasia-optic nerve decussation defect-anterior segment dysgenesis syndrome. This is because both descriptions mention foveal hypoplasia, optic nerve misrouting or decussation defects, and anterior segment dysgenesis, all in the absence of albinism. Additionally, both conditions present with congenital nystagmus, decreased visual acuity, and refractive errors.
truth:  Foveal hypoplasia-optic nerve decussation defect-anterior segment dysgenesis syndrome

\paragraph{LLaMa:}
It depends on your perspective, but I would say that all of these options could be considered the same concept as "foveal hypoplasia 2" since they all have some common symptoms and characteristics. For example:

Option 0 mentions "anterior segment dysgenesis," which is a type of eye abnormality that can affect the development of the cornea, iris, or lens. This is similar to "anterior segment dysgenesis" mentioned in your original description of "foveal hypoplasia 2."
Option 1 mentions "optic nerve hypoplasia/aplasia," which means that the optic nerve is underdeveloped or missing entirely. This is similar to the "optic nerve decussation defects" mentioned in your original description of "foveal hypoplasia 2."
Option 2 mentions "microphthalmia-retinitis pigmentosa-foveoschisis-optic disc drusen syndrome," which is a rare genetic disorder that can cause small eyes (microphthalmia), a breakdown of the retina (retinitis p
Foveal hypoplasia-optic nerve decussation defect-anterior segment dysgenesis syndrome
(cut off becasue of max output limit)

\paragraph{Case6:}
hypermanganesemia with dystonia 2  ||  Dystonia-parkinsonism-hypermanganesemia syndrome
\paragraph{GPT4:}
The same concept to "hypermanganesemia with dystonia 2" is option 2: Cirrhosis-dystonia-polycythemia-hypermanganesemia syndrome. This is because it is also a rare disorder of manganese transport characterized by the presence of dystonia and hypermanganesemia, along with other symptoms such as liver cirrhosis and polycythemia. The brain imaging findings in this syndrome are also consistent with the deposition of manganese in the basal ganglia and other regions, similar to "hypermanganesemia with dystonia 2".
LLama:
I'm sorry, but none of those options are exactly the same as "hypermanganesemia with dystonia 2." However, one of them does come close. Option 0, Dystonia-parkinsonism-hypermanganesemia syndrome, also involves hypermanganesemia and dystonia, although it may have other symptoms as well.
        
        The main difference between "hypermanganesemia with dystonia 2" and Dystonia-parkinsonism-hypermanganesemia syndrome is that the former seems to be a specific type of the latter. In other words, all people with "hypermanganesemia with dystonia 2" would have Dystonia-parkinsonism-hypermanganesemia syndrome, but not everyone with Dystonia-parkinsonism-hypermanganesemia syndrome would have "hypermanganesemia with dystonia 2."
        
        I hope that helps! Let me know if you have any other questions.

\paragraph{Case7:}
epiphyseal dysplasia, multiple, with miniepiphyses  ||  Multiple epiphyseal dysplasia, with miniepiphyses

\paragraph{GPT4:}
The same concept to "epiphyseal dysplasia, multiple, with miniepiphyses" is option 0: Multiple epiphyseal dysplasia, with miniepiphyses. This is because it describes the same disorder characterized by strikingly small secondary ossification centers (mini-epiphyses) in all or only some joints, resulting in severe bone dysplasia of the proximal femoral heads, along with short stature, increased lumbar lordosis, genua vara, and generalized joint laxity.

\paragraph{LLaMa:}
The correct answer is option number one: Multiple epiphyseal dysplasia, with miniepiphyses.

Multiple epiphyseal dysplasia (MED) is a group of rare skeletal disorders characterized by abnormalities in the growth and development of the bones and joints. MED can be divided into two main types based on their genetic cause: autosomal dominant MED, which is caused by mutations in the COMP gene, and recessive MED, which is caused by mutations in one of several other genes.

One of the distinguishing features of MED is the presence of miniepiphyses, which are small bony projections that develop at the ends of long bones instead of the normal single growth plate (epiphysis). These miniepiphyses can cause abnormalities in the shape and alignment of the bones, leading to joint pain, limb deformity, and early onset of osteoarthritis.

The term "multiple epiphyseal dysplasia, with miniepiphyses" is used specifically to describe a form of
Multiple epiphyseal dysplasia, with miniepiphyses(cut off becasue of max output limit)

\paragraph{Case8:}
myopathy, myofibrillar, 3  ||  Distal myotilinopathy

\paragraph{GPT4:}
The same concept to "myopathy, myofibrillar, 3" is 1: Myofibrillar myopathy. This is because both terms refer to a group of skeletal and cardiac muscle disorders characterized by the disintegration of myofibrils and aggregation of degradation products into intracellular inclusions. They also share similar clinical features, such as slowly-progressive muscle weakness that can affect proximal, distal, cardiac, and respiratory muscles.

\paragraph{LLaMa:}

code

\paragraph{Case9:}
2-methylbutyryl-coa dehydrogenase deficiency  ||  2-methylbutyryl-CoA dehydrogenase deficiency

\paragraph{GPT4:}
The same concept to "2-methylbutyryl-CoA dehydrogenase deficiency" is option 0: 2-methylbutyryl-CoA dehydrogenase deficiency. This is because it has the same name and refers to the same condition, which is characterized by impaired isoleucine degradation with increased plasma or whole blood C5 acylcarnitine levels and increased urinary excretion of N-methylbutyrylglycine. The condition is usually clinically asymptomatic, but some patients may experience muscular hypotonia, developmental delay, and seizures.

\paragraph{LLaMa:}
option 0.It's because they have one or more related keywords in common.
        
        Here are the keywords of 2-methylbutyryl-CoA dehydrogenase deficiency:
            - "2-methylbutyryl glycinuria"
            - "short/branched-chain acyl-coa dehydrogenase deficiency"
        
        And here are the keywords of Isobutyryl-CoA dehydrogenase deficiency:
            - "2-methylbutyryl glycinuria"
            - "short/branched-chain acyl-coa dehydrogenase deficiency"
        
        As you can see, they have the same keywords in common. So it's reasonable to say that they are related.
        
        The other options don't have these specific keywords, so they are less likely to be related.







\end{document}